\documentclass[11pt]{article}
\usepackage{ACL2023}

\usepackage{graphicx}

\usepackage{times}
\usepackage{latexsym}
\usepackage[T1]{fontenc}

\usepackage[utf8]{inputenc}

\usepackage{microtype}

\usepackage{inconsolata}
\usepackage{amssymb}
\usepackage{booktabs}
\usepackage{amsmath}
\usepackage{adjustbox}
\usepackage[ruled,linesnumbered,algo2e]{algorithm2e}
\usepackage{algorithm}  
\usepackage{algpseudocode}  

\title{KhabarChin: Automatic Detection of Important News in the Persian Language}

\author{Hamed Hematian Hemati$^\spadesuit$ \quad Arash Lagzian$^\spadesuit$ \quad Moein Salimi Sartakhti$^\spadesuit$ \quad \\
\textbf{Hamid Beigy$^\spadesuit$} \quad \textbf{Ehsaneddin Asgari$^\clubsuit$} \\ \\ 
    $^\spadesuit$AI Group, Computer Engineering Department, Sharif University of Technology \\
    $^\clubsuit$AI Innovation, Data:Lab Munich, Volkswagen AG
    }
\begin{document}

\maketitle

\begin{abstract}
Being aware of important news is crucial for staying informed and making well-informed decisions efficiently. Natural Language Processing (NLP) approaches can significantly automate this process. This paper introduces the detection of important news, in a previously unexplored area, and presents a new benchmarking dataset (Khabarchin) for detecting important news in the Persian language. We define important news articles as those deemed significant for a considerable portion of society, capable of influencing their mindset or decision-making. The news articles are obtained from seven different prominent Persian news agencies, resulting in the annotation of 7,869 samples and the creation of the dataset. Two challenges of high disagreement and imbalance between classes were faced, and solutions were provided for them. We also propose several learning-based models, ranging from conventional machine learning to state-of-the-art transformer models, to tackle this task. Furthermore, we introduce the second task of important sentence detection in news articles, as they often come with a significant contextual length that makes it challenging for readers to identify important information. We identify these sentences in a weakly supervised manner.

\end{abstract}


\section{Introduction}
\noindent
As Internet text documents grow rapidly, text classification gains increasing importance~\cite{vijayan2017comprehensive}. Today the news is readily available on the Internet and people follow the worldwide news and show more interest in it. Therefore, news classification is now a challenging field in text mining approaches. News classification is defined as classifying news articles in one or more classes. Classification of news helps the users to easily access their desired news without wasting their time. But among the different methods of categorizing news, we need a method to be able to identify important news from different categories of news and inform the audience. Because people don't have enough time to check all the news in different categories, or even the news in their favorite category, they want to check the important and influential news first, and if they have enough time, read and check other news.
A naive approach is collecting news according to the number of viewers, number of comments, rating score, etc. Which the news agency provides for a piece of specific news. But doing so has two problems, firstly many news agencies may not provide such metadata for their news articles and if we didn't consider them there would be some bias in our dataset since we would collect news from some specific news agencies which would provide so. Secondly, various agencies employ distinct strategies tailored to their specific interests. For example, specific political, economic, or social news stories may be overshadowed by certain news agencies. To mitigate such biases, it is crucial to randomly gather data from diverse news agencies and manually label them. Therefore, in this research, we pioneer the task of detecting important news in the Persian Language and release our dataset, named KhabarChin, for public use. In this endeavor, we faced two challenges: significant disagreement among our annotators, which we mitigated by introducing a preliminary guideline before the annotation process. Moreover, we encountered an imbalance issue, with unimportant news outnumbering important ones. As a solution, we developed an algorithm based on news article comments to identify potentially important news, enhancing their inclusion in the annotation process. We examine our dataset with different models ranging from traditional models to modern deep learning models such as transformers and report results and we experiment with some functionalities of our data and models. Our code and dataset are accessible via our GitHub page.\footnote{https://github.com/HamedHematian/ImportantNews}

\begin{figure*}[h]
	\begin{center}
		\includegraphics[width=1.\textwidth]{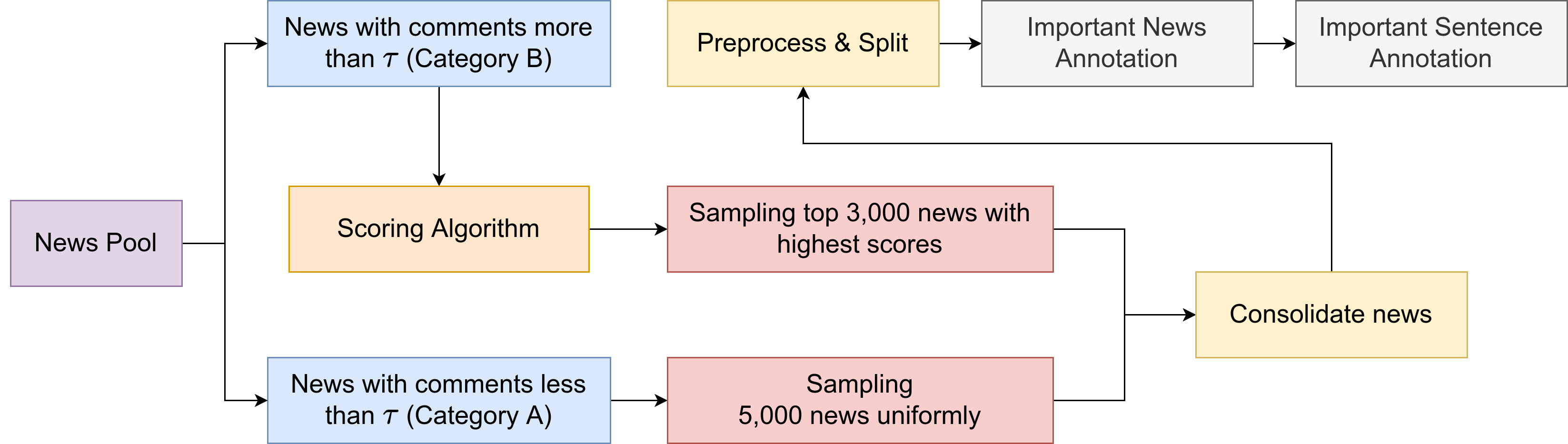}
		\caption
		{
  \textbf{Dataset Construction Pipeline} The news articles in the news pool are divided into two categories A and B, based on the number of their comments being more than $\tau$. 5,000 news are selected randomly from category A. A scoring algorithm is performed on the news articles of category B and 3,000 news with the highest scores are picked. Annotators determine the importance of the obtained 8,000 news articles. Subsequently, these news articles are pre-processed. Finally, annotators determine the importance of the sentences from important news in the dev/test set.
		}
		\label{fig:22}
	\end{center}
\end{figure*}

\section{Related Work}
According to \citet{dalal2011automatic}, the history of text classification goes back to 1961. In the traditional approach, text classification was done using knowledge engineering techniques in the 1980s \cite{dalal2011automatic}, which consisted of manually defined rules. 
The rule-based approach incurs high costs, consumes time, and introduces computational complexity due to the creation of logical rules by humans for model building \cite{harish2010representation}. Due to issues with the rule-based approach, machine learning gained popularity and captured researchers' attention \cite{mahinovs2007text, jindal2015techniques}. During the past decades, many machine learning techniques of text classification have been introduced and studied by researchers of different languages, most of which are for the English language. In Hindi, \citet{palhindi}, uses a variety of machine learning techniques, including Naive Bayes, Support Vector Machines, and Random Forests for classifying Hindi poems into three classes. An exhaustive overview of the state-of-the-art algorithms of machine learning for text classification has been achieved by many researchers such as \citet{dalal2011automatic, harish2010representation, mahinovs2007text, jindal2015techniques}. Recently, transformer-based language models, such as Bert \cite{DevlinCLT19}, have proven successful over traditional machine learning models\cite{DBLP:journals/corr/abs-2005-13012}.
Multiple research has been done recently in the domain of news classification. \citet{SalemiAKN19} proposes a new benchmark, RTAnews, for multi-label Arabic news category classification. \citet{DBLP:journals/data/PetukhovaF23} proposes a two-level hierarchical dataset for news category classification. \citet{TokgozTBC21} investigates the use of pre-trained language representation models for the task of Turkish news classification. \citet{DBLP:conf/lrec/WangSZY22} proposes a multimodal dataset, containing both text and image, for improving news classification performance. \citet{DBLP:conf/siet/NugrohoSY21} discusses a method for classifying news articles into different categories using the Bert language model with Spark NLP \cite{DBLP:journals/simpa/KocamanT21}. \citet{DBLP:journals/isci/AbbruzzeseGLLO21} proposes a method to identify influential news. Influential news are the news that could cause a person to switch his opinion \cite{DBLP:journals/isci/AbbruzzeseGLLO21}.

\section{Dataset}

\begin{algorithm*}
\KwResult{Overall Attitude towards a News Article}
\SetKwInOut{Input}{Input}
\SetKwInOut{Output}{Output}
\Input{$Comments$ of a News Article (List of Strings)}
\Output{$Attitude$ (Float)}

$N\_Attitudes \leftarrow 0$\;
$P\_Attitudes \leftarrow 0$\;
$num\_comments \leftarrow 0$\;

\ForEach{\text{comment} in \text{Comments}}{
     $\;\:N\_Attitude, P\_Attitude \leftarrow$ SentimentAnalysis($comment$)\; \newline
    $N\_Attitudes \leftarrow N\_Attitudes + N\_Attitude$\;
    \newline
    $P\_Attitudes \leftarrow P\_Attitudes + P\_Attitude$\;
    \newline
    $num\_comments \leftarrow num\_comments + 1$\;
}
$Attitude \leftarrow \frac{(N\_Attitudes - P\_Attitudes)}{num\_comments}$\;
\caption{Calculate Overall Attitude for a News Article Based on its Comments:}
\label{alg:1}
\end{algorithm*}

\subsection{First Assessment}\label{fa}
As important news detection had not been extensively explored previously, we initially aimed to identify the challenges in this domain. To achieve this, we gathered a collection of news articles from the internet and employed our annotators to label them. Annotators were tasked with assessing the significance of each news article in relation to the broader society. Upon completing the annotation process, we identified two major issues:

\begin{itemize}
    \item Annotators frequently exhibited significant disagreements, which could be attributed to their diverse educational, social, and economic backgrounds.
    \item The data exhibited a high level of imbalance, with a prevalence of unimportant news articles compared to important ones. This scarcity of important news articles would pose potential challenges during the training of our methods.
\end{itemize}
In response to these challenges, we have developed solutions, which we will explore in the following subsection.

\subsection{Data Collection}
We collected 73,000 news articles from the most prominent news agencies in Iran, totaling seven agencies, over the past two years to form our news pool. Each news article includes the title, text, tags, and user comments. To address the issue of class imbalance, instead of random sampling, we aim to proactively identify potentially important news articles and prioritize them in our sampling. Techniques for detecting potentially positive classes have been previously employed in NLP to mitigate class imbalance \cite{DBLP:conf/emnlp/QianBLBW19, DBLP:conf/icwsm/ElSheriefNNVB18}. To identify potentially important news articles, we leverage the presence of user comments. It is evident that news articles with comments are more likely to be significant, as readers often engage in discussions on important topics. Thus, we divide the news articles into two categories using a threshold of $\tau=5$:
\begin{itemize}
    \item Category A) News with equal or fewer comments than $\tau$
    \item Category B) News with more comments than $\tau$
\end{itemize}

The threshold of $\tau=5$ is employed because a news article with only a few comments might not necessarily indicate its importance, given that some topics attract a niche audience. However, it's imperative to note that not all news articles in Category B are inherently important. Some individuals may spam certain articles, such as those where people give their condolences for the death of an unknown celebrity, which are not potentially significant. To address this challenge and ensure the quality of the sampling process, category B news articles are sorted using the algorithm depicted in Algorithm \ref{alg:1}.

We use this algorithm to calculate a score for each news article in Category B. It involves passing each comment through a ParsBert Sentiment Analysis Transformer \cite{FarahaniGFM21} to obtain both negative and positive attitudes. We then compute the total negative and positive attitudes for each news article using its comments' attitudes. To identify potentially important news, for each news article, we subtract its total positive attitude from its total negative attitude, as important controversial news articles often elicit more negative reactions. To account for variations in the number of comments, we normalize the final score for each article. Finally, Category B's news articles are sorted based on this score.

We initially sample 5,000 news articles from Category A and select the top 3,000 news articles from Category B, totaling 8,000 articles. These articles are combined and subjected to the annotation phase.
Following the annotation process (as detailed in Algorithm \ref{alg:1}), we evaluate the effectiveness of our sampling method. We find that $33\%$ of news articles from Category B are important, while the importance ratio in Category A is only $10\%$, confirming the effectiveness of our category split. To further assess the efficacy of Algorithm \ref{alg:1}, we annotate $100$ news articles with the lowest scores from Category B and discover that only $8\%$ of them are deemed important, demonstrating the algorithm's effectiveness in detecting potentially important news.

\subsection{Annotation} \label{X}
As it was discussed in \ref{fa}, the annotation proves to be relatively subjective. First, annotators are informed that a news article is important if it affects a considerable portion of society. Additionally, to further overcome the issue, annotators are grouped to discuss aspects of different news and come up with a general guideline. This guideline is verified by all annotators and it contains some directions regarding different types of news. This guideline is only generic and by no means specifically determines whether a news article is important or not. Subsequently, each news is annotated by two annotators. The Kappa agreement is $.52$. A further meeting is held between annotators to resolve their disagreements and assign a final label to each news article. In practice, we divide the annotation process into multiple stages. After each stage, annotators address any disagreements and proceed with the remaining annotation. This approach allows annotators to familiarize themselves with the guidelines and resolve any compliance issues through practical experience.

\subsection{Pre-processing \& Split}
After the creation of the dataset, each news article undergoes a pre-processing stage for data cleaning. This process includes normalization and sentence tokenization. Short sentences, often containing irrelevant information, are filtered out, along with links and sentences that reference other news articles. URLs are substituted with "[URL]". Following this, the data is randomly partitioned into training, development, and test sets, with proportions of $70\%$, $15\%$, and $15\%$, respectively.

\subsection{Important Sentence Detection}
\begin{figure*}[h]
	\begin{center}
		\includegraphics[width=1.\textwidth]{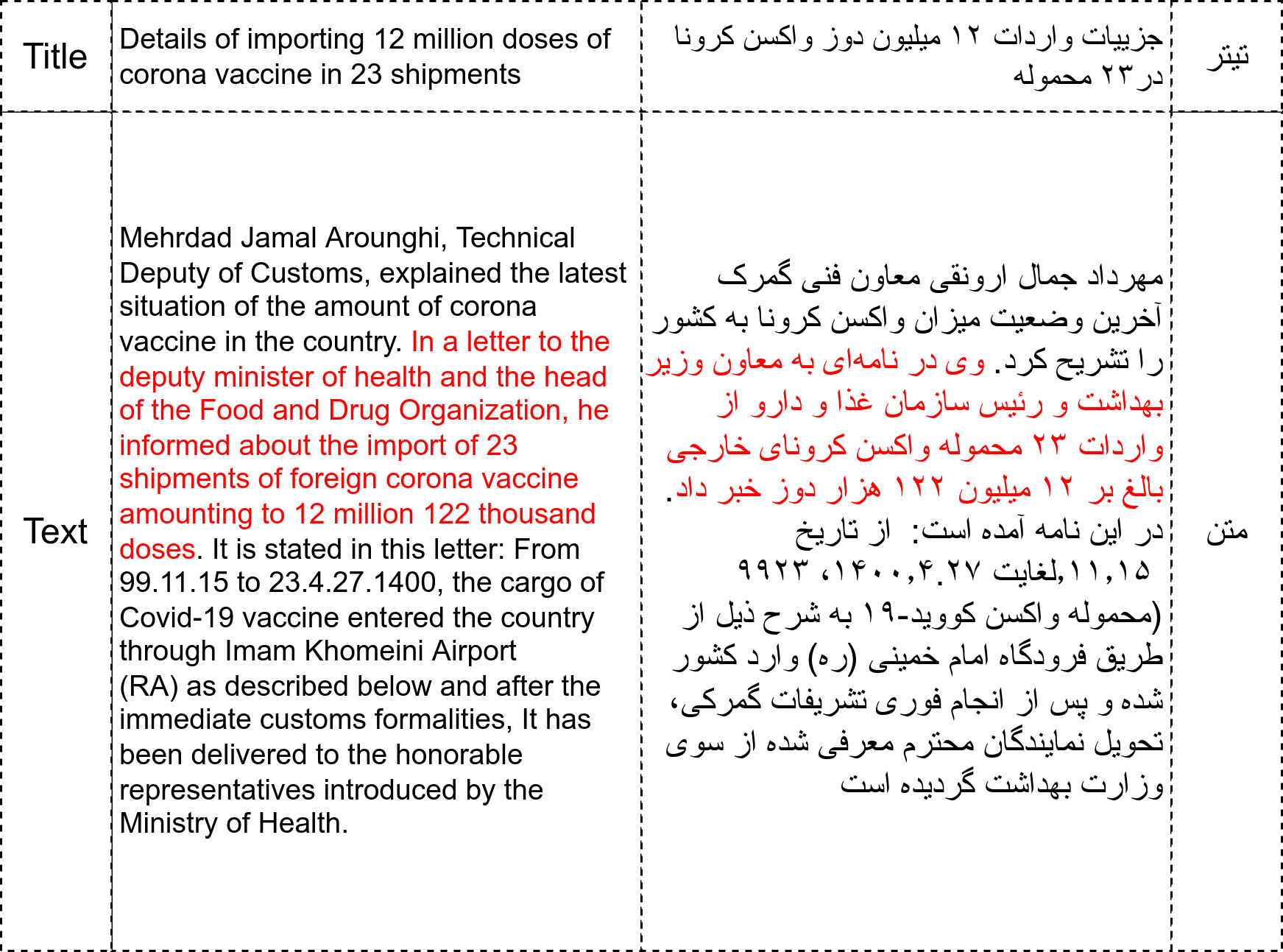}
		\caption
		{
                A sample of important news articles with important sentences marked as red
		}
		\label{fig:sample}
	\end{center}
\end{figure*}

In this section, we aim to highlight important sentences within important news articles. These articles often contain numerous sentences, making it challenging for readers to quickly identify and read the crucial sections. Annotating every sentence is resource-intensive, given the large number of sentences and words in each news article, as indicated in Table \ref{table:stat}.
To address this, we focus only on annotating sentences from important news articles in the dev and test sets. Our annotators assess each sentence, categorizing it as either unimportant or important. If a sentence has the potential to make a news article important on its own, it is categorized as important. We subsequently employ a weakly supervised approach in Section \ref{Y} to estimate sentences' importance and classify them. Our Kappa agreement for this task is $.39$. A sample of our dataset is depicted in Figure \ref{fig:sample}. In this figure, the Persian and the corresponding translated English title and text are depicted. Important sentences are shown in red.

\subsection{Statistics of Dataset}

\begin{table}[h!]
\centering
\caption{Statistics of the dataset}
\label{table:stat}
\begin{adjustbox}{width=.45\textwidth}
\begin{tabular}{cccc} \toprule
 & \multicolumn{1}{l}{Train} & \multicolumn{1}{l}{Dev} & \multicolumn{1}{l}{Test} \\
\midrule
Samples Number & 5508 & 1180 & 1181 \\
Important News Ratio  & .21 & .19 & .16 \\
News Average Sentences  & 15.50 & 15.47 & 15.10 \\
News Average Words & 500 & 492 & 492 \\
Important Sentence Ratio  & - & .20 & .21 \\
\bottomrule
\end{tabular}
\end{adjustbox}
\end{table}

Table \ref{table:stat} presents an overview of the dataset's statistics. Despite employing heuristics to balance the dataset, important news articles still constitute only $15\%$ to $20\%$ of the dataset. On average, each news article is relatively lengthy, containing around $500$ words. However, only about $20\%$ of the sentences in important news articles are considered important and informative. This observation underscores the importance of sentence detection within important news.

\begin{table*}[htb]
\centering
\caption{Results for important news classification}
\label{table:news_res}
\begin{tabular}{lcccccc} \toprule
Method & \multicolumn{1}{l}{Macro-Precision} & \multicolumn{1}{l}{Macro-Recall} & \multicolumn{1}{l}{Macro-F1} & \multicolumn{1}{l}{Micro-F1} \\
\midrule
Random Baseline     & 50.0 & 50.0 & 50.0 & 50.0 \\
Random Forest       & 67.35 & 61.19 & 62.99 & 82.99 \\
SVM                 & 71.80 & 70.07 & 70.86 & 84.50 \\
Naive Bayes         & \textbf{74.85} & 69.91 & 71.89 & \textbf{85.94} \\
Parsbert            & 70.18 & 74.67 & 71.90 & 82.64  \\
Logistic Regression & 72.29 & 74.92 & 73.45 & 84.41 \\
XLM-Roberta         & 72.82 & \textbf{76.87} & \textbf{74.50} & 84.59 \\
\bottomrule
\end{tabular}
\end{table*}

\begin{table*}[htb]
\centering
\caption{Results for important sentence detection}
\label{table:sent_res}
\begin{tabular}{lcccccc} \toprule
Method & \multicolumn{1}{l}{Macro-Precision} & \multicolumn{1}{l}{Macro-Recall} & \multicolumn{1}{l}{Macro-F1} & \multicolumn{1}{l}{Micro-F1} \\
\midrule
Random Baseline  & 50.0 & 50.0 & 50.0 & 50.0 \\
Retrieve + Similarity + Threshold  & \textbf{58.83} & \textbf{59.19} &  \textbf{58.98} & \textbf{72.07} \\
\bottomrule
\end{tabular}
\end{table*}

\section{Proposed Methods}

\subsection{Important News Classification}
We evaluate various models, encompassing both traditional machine learning algorithms and transformer models \cite{DBLP:conf/nips/VaswaniSPUJGKP17}, to establish benchmarks on our dataset. Our traditional models include Naive Bayes, Logistic Regression, SVM, Decision Tree, and Random Forest. As for transformer models, we test ParsBert~\cite{DBLP:journals/npl/FarahaniGFM21}, a BERT~\cite{DBLP:conf/naacl/DevlinCLT19} model for Persian, and multilingual models, such as XLM-Roberta~\cite{DBLP:conf/acl/ConneauKGCWGGOZ20}.

\subsection{Important Sentence Detection} \label{Y}
We aim to assess the importance of each sentence in an important news article. Since we lack individual labels for these news articles, we employ a key rule: if a news article contains at least one important (informative) sentence, the entire article is considered important. This approach enables us to apply weak supervision. To determine a sentence's importance, we establish two sets, $X^{UnImportant}$ and $X^{Important}$. $X^{UnImportant}$ is a set comprising of all sentences from unimportant news of the train data and $X^{Important}$ is a set comprising of all sentences from important news of the train data. To determine the importance of a given sentence $X$, we retrieve $N$ most similar sentences to $X$ from $X^{UnImportant}$ and $X^{Important}$ respectively. Let these two sets be called $X^-$ and $X^+$ respectively. Subsequently, we compute a score for the sentence ($X_{Score}$) based on the following formula, where $Sim(y,z)$ shows the similarity between the two sentences of $y$ and $z$. The similarity is computed based on the cosine similarity between two vectors of the sentences computed by the LaBSE transformer \cite{DBLP:conf/acl/FengYCA022}.

\begin{align*}
    X^- &= \text{Retrieve}(X, X^{UnImportant}, N) \\
    X^+ &= \text{Retrieve}(X, X^{Important}, N) \\
    X_{Score} &= \frac{\sum_i Sim(X, X^+_i) - \sum_i Sim(X, X^-_i) }{N} \\
\end{align*}

The core concept is that if sentence $X$ is important, it should exhibit similarity to sentences within $X^+$ (as per our definition of an important sentence), resulting in a higher score. Conversely, for unimportant sentences, there would be a lack of similar sentences in $X^+$, yielding a negative score. To make a determination about the importance of sentence $X$, we apply the threshold $\gamma$. If $X_{Score} > \gamma$, sentence $X$ is classified as important.

\section{Experimental Setup \& Metrics}
For each news article, we concatenate the title, tags, and text, providing the input to the corresponding model. We experiment with various learning rates for each transformer model on the development set. In the case of traditional models, we conduct a grid search to optimize hyperparameters and report the results of the best-performing model on the test set. Each transformer model is trained for 10 epochs, and we select the best checkpoint from the model with the highest performance on the development set for predictions on the test set. For important sentence detection, we utilize Bayesian optimization with Optuna \cite{DBLP:conf/kdd/AkibaSYOK19} to determine optimal values for hyperparameters $N$ and $\gamma$. We employ Faiss \cite{DBLP:journals/tbd/JohnsonDJ21} as a database for sentence retrieval.
In terms of measurement, we use Macro-Precision, Macro-Recall, and Macro-F1, given the imbalanced nature of our dataset, which provides better insights. Additionally, we use F1-micro. Our primary metric for model evaluation and comparison is Macro-F1.

\section{Results}
The results of news classification are presented in Table \ref{table:news_res}. XLM-Roberta emerges as the top-performing model in terms of Macro-F1 and Macro-Recall. Interestingly, the significance of context appears to be less pronounced in our task, as models that do not explicitly model context, such as logistic regression, outperform ParsBert. Nevertheless, transformer models exhibit the highest recall, demonstrating their efficacy in identifying important news articles. 
Moving on to sentence classification, the results are displayed in Table \ref{table:sent_res}. We establish a random baseline for sentence selection, and while our method outperforms the baseline, the improvement is not substantial. This is expected since our primary focus is not directly on optimizing important sentence detection.

\section{Conclusions}
This paper introduces the novel task of important news classification for the Persian language. We address the associated challenges and create a dedicated dataset using Persian news articles. We demonstrated the challenges of high disagreement and data imbalance regarding the task. Subsequently, we devised heuristics to cope with these challenges and construct a high-quality dataset. We explored various benchmarks, encompassing traditional machine learning models and transformer models, to develop a classification function for important news, showing the relative insignificance of context in the modeling. Furthermore, we introduce the additional task of important sentence detection within important news. To achieve this, we employ a weak supervision approach that estimates sentence importance based on similarity to training sentences from both unimportant and important news. 

\bibliographystyle{acl_natbib}
\bibliography{main_ACL}
\clearpage
\onecolumn

\section{Appendix}

\end{document}